\documentclass[runningheads]{llncs}
\usepackage{times}
\usepackage{epsfig}
\usepackage{graphicx}
\usepackage{amsmath}
\usepackage{amssymb}
\usepackage{graphicx}
\usepackage{subcaption}

\begin{document}
\title{Causal Scoring Medical Image Explanations: \\ 
A Case Study On Ex-vivo Kidney Stone Images 
}
\titlerunning{Causal Scoring Medical Image Explanations}  

\author{Armando Villegas-Jimenez$^{1}$
\and
Daniel Flores-Araiza$^{2}$
\and
Francisco Lopez-Tiro$^{2,3}$
\and
Gilberto Ochoa-Ruiz$^{2}$
\and
Christian Daul$^{3}$
%
%
}
\authorrunning{Villegas-Jimenez, A. et al.}  

\institute{
Instituto Politécnico Nacional, Mexico \and
Tecnologico de Monterrey, School of Engineering and Sciences, Mexico \and
CRAN UMR 7039, Université de Lorraine and CNRS, Nancy, France
}


\maketitle

\begin{abstract}
On the promise that if human users know the cause of an output, it would enable them to grasp the process responsible for the output, and hence provide understanding, many explainable methods have been proposed to indicate the cause for the output of a model based on its input. 
Nonetheless, little has been reported on quantitative measurements of such causal relationships between the inputs, the explanations, and the outputs of a model, leaving the assessment to the user, independent of his level of expertise in the subject.
To address this situation, we explore a technique for measuring the causal relationship between the features from the area of the object of interest in the images of a class and the output of a classifier. 
Our experiments indicate improvement in the causal relationships measured when the area of the object of interest per class is indicated by a mask from an explainable method than when it is indicated by human annotators.
Hence the chosen name of \textit{Causal Explanation Score (CaES)}

\end{abstract}

\section{Introduction}

The early identification of the type of kidney stone relies on the need for such classification by a urologist to determine and begin treatment.
Also, several developed countries, as noted by \cite{kasidas_renal_2004} and \cite{hall_nephrolithiasis_2009}, report a significant incidence of urinary lithiasis (formation or presence of kidney stones), with around 10\% of individuals experiencing a kidney stone episode at least once. 
Furthermore, there's a notably high recurrence rate of 40\% in these countries.
Commonly, the analysis and classification procedure for kidney stones, known as Morpho-Constitutional Analysis (MCA) \cite{CORRALES202113} is time-consuming, expensive and requires a great deal of experience. 
Moreover, medical image analysis, as well as MCA, have been demonstrated to be highly operator dependent \cite{de_coninck_metabolic_2019,Ultrasound_Operator_dependent_2022,aakesson_operator_dependent2004}.
Additionally to these issues, due to the increasing number of patients each year and given the naturally high diversity of medical cases, the realm of medicine is perpetually in need of methods that are both more accurate and faster  \cite{topol_deep_2019}.

\begin{figure}[t] 
\centering
    \includegraphics[width=1.0\linewidth]{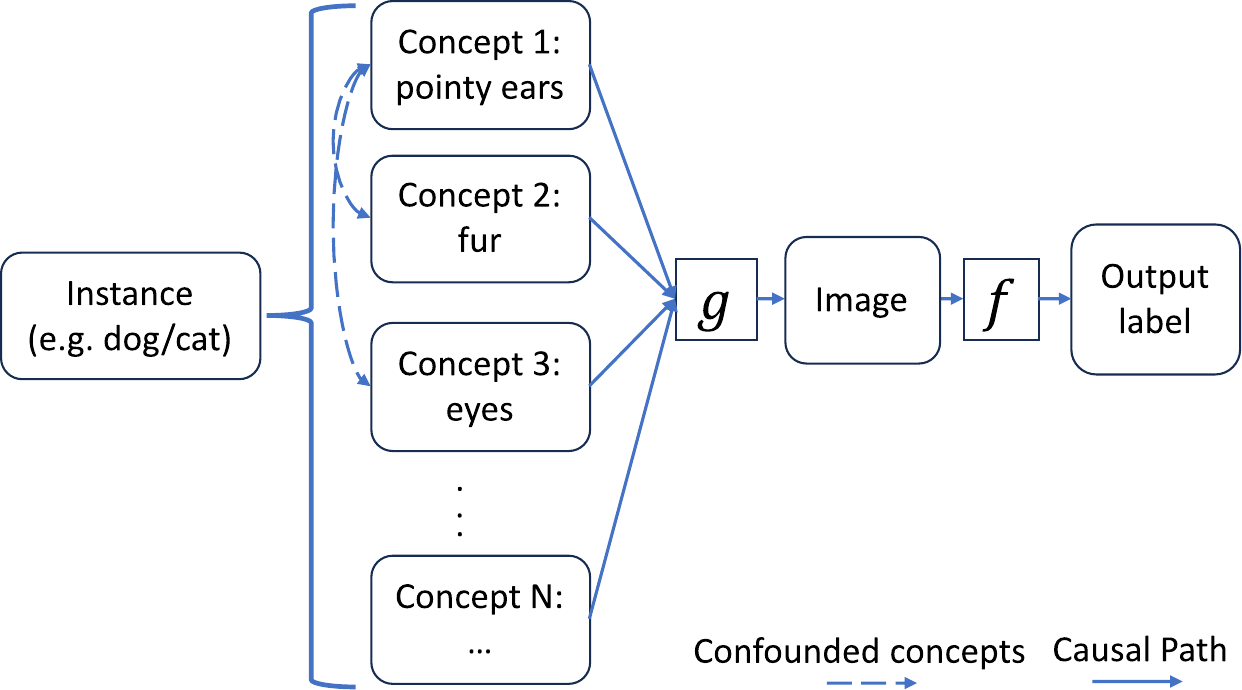}
    \caption{\small{
    Causal graph relating instances to high-level concepts (such as cat to fur, pointy ears, etc.), images and a Deep Learning classifier $f$, which gives an output label. 
    The dashed edge indicates possible confounding between concepts.  
    Edges connecting the concepts to the image, through $g$, correspond to the natural generation process of images.
    }} 
\label{fig_causal_concept_diagram}
\end{figure}

\begin{figure*}[t]
\centering
    \includegraphics[width=1.0\linewidth]{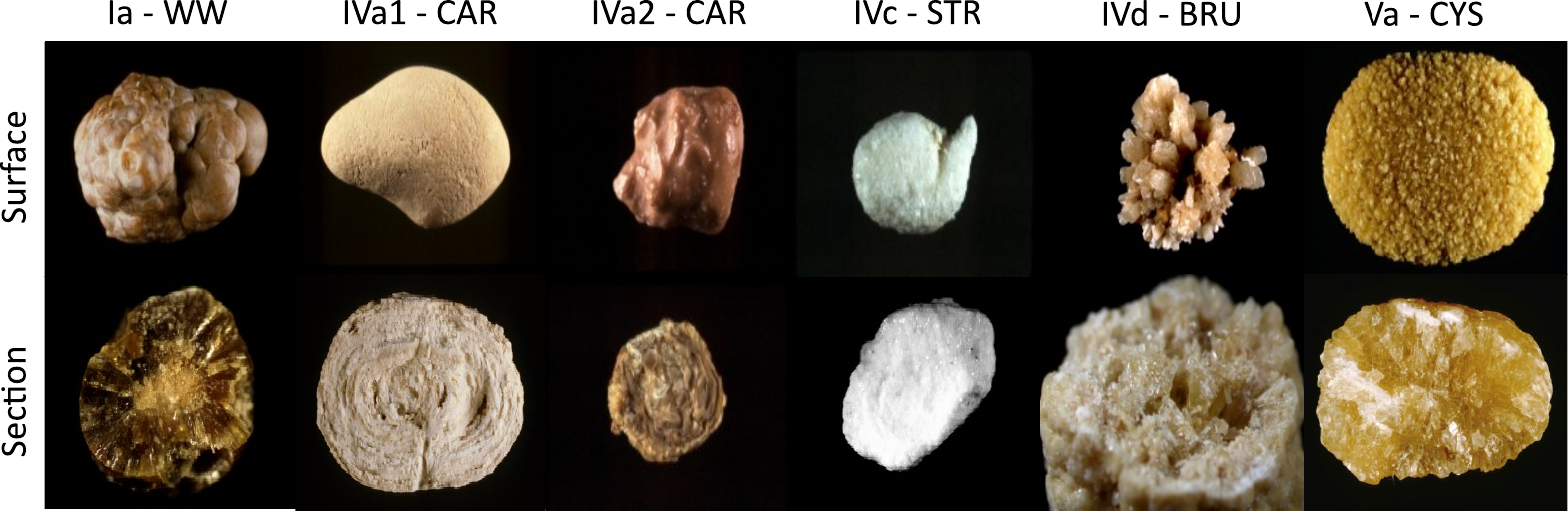} 
    \caption{\small{Examples of the six different sub-types of kidney stones from the dataset \cite{CORRALES202113}. The complete name of those stones are Whewellite (Ia - WW), Carbapatite (IVa1 \& IVa2, CAR), Struvite (IVc - STR),  Brushite (IVd -BRU), and Cystine (Va - CYS). Here we show examples of both types of views in the dataset per class, images of the ``Surface" of the stones, and cross ``Section" views. }} 
\label{fig_dataset}
\end{figure*}

Recent strides in the field of AI, specifically advances in Deep Learning (DL) have boosted the performance and instigated the early adoption of DL models in medical imaging \cite{lee_current_2019,jiang_review_2023}.
In the context of MCA, methods developed for in-vivo or ex-vivo kidney stone recognition \cite{ochoa-ruiz2022vivo,lopez2021assessing,gonzalez_zapata2023metric,zhu2023segprompt} have been proposed. 
Although most DL-based methods outperform non-DL methods in terms of accuracy, they lack explainability capabilities as  DL models simply output a classification for an input regardless if it is even adequate or not. 
However, since the field of medical image analysis brings with it high-stakes decisions, chiefly the diagnosis, which carries a direct and profound impact on patient's lives, the need for robust automated analysis of medical images for Computer-Aided Diagnosis (CADx) cannot be overstated \cite{Rudin2019stop}. 
Hence, medical specialists need to understand how features on the input image caused the output of the DL model \cite{Flores_Araiza_2023_CVPR}.
Precisely, the field of eXplainable AI (XAI) seeks to provide an understanding of the behavior of a DL model.
Under this primary objective, most of the XAI methods proposed in the literature highlight the \textit{causes} of the output of a model from its input \cite{borys_explainable_2023}, 
i.e. explanations attempt to reflect the natural generation process $g$ depicted on Fig.\ref{fig_causal_concept_diagram}. 
Nonetheless, this causal relationship assumed to be present in the explanations has been left without a quantitative measure.

Therefore, in order to address such shortcomings, in this work we have 
\textit{I)} adapted a method, in an ex-vivo kidney stone dataset \cite{CORRALES202113}, for causal scoring the relationship between the latent features in an image classifier and its output classification \cite{Causal_signals_LopezPaz_2017_CVPR}. 
This adaptation enables a more precise measurement by working with segmentation masks, instead of bounding boxes. Moreover, our adaptation outputs a causal measurement between 0 and 1, instead of any positive value. 
Additionally,  
\textit{II)} we modified an explainable AI method (GradCAM \cite{GradCAM_Ramprasaath_2016}), to automate extracting binary segmentation masks of the region of highest interest for the model, allowing to obtain more consistent causal measurements than with human-annotated segmentation masks.

Our method is named Causal Explanation Score (CaES).
With CaES, we provide a method to causally validate the outputs of a DL model based on the areas in the input image indicated by the explanations generated for the same input and output of the model. 
More importantly, we show results indicating our method (CaES) attains better results than when using human-annotated segmentation masks of the objects of interest.
Our work is thus focused on enabling healthcare specialists to leverage DL model findings for diagnosis, grasping the logic behind such results, while safeguarding the responsible application of these powerful AI technologies in medical diagnostics, striking a crucial balance between machine efficiency and human accountability. 

%

\section{Dataset and Methods}

\subsection{Kidney stones dataset}\label{KS_dataset}

Our ex-vivo dataset, Fig.\ref{fig_dataset}, is split into 209 surface and 157 section images, 366 in total.
Those images were acquired with a digital camera (CCD) under controlled lighting conditions and with a uniform background.
The dataset is sorted by the kidney stones sub-types, six in our case. 
Those subtypes, as shown in Fig.\ref{fig_dataset},  
 are \textit{Whewellite, sub-type Ia (Ia - WW), Carbapatite sub-type IVa1 (IVa1 - CAR), Carbapatite sub-type IVa2 (IVa2 - CAR), Struvite sub-type IVc (IVc - STR), Brushite sub-type IVd (IVd - BRU) and Cystine sub-type Va (Va - CYS)}. 

\begin{figure*}[t] 
\centering
    \includegraphics[width=0.75\linewidth]{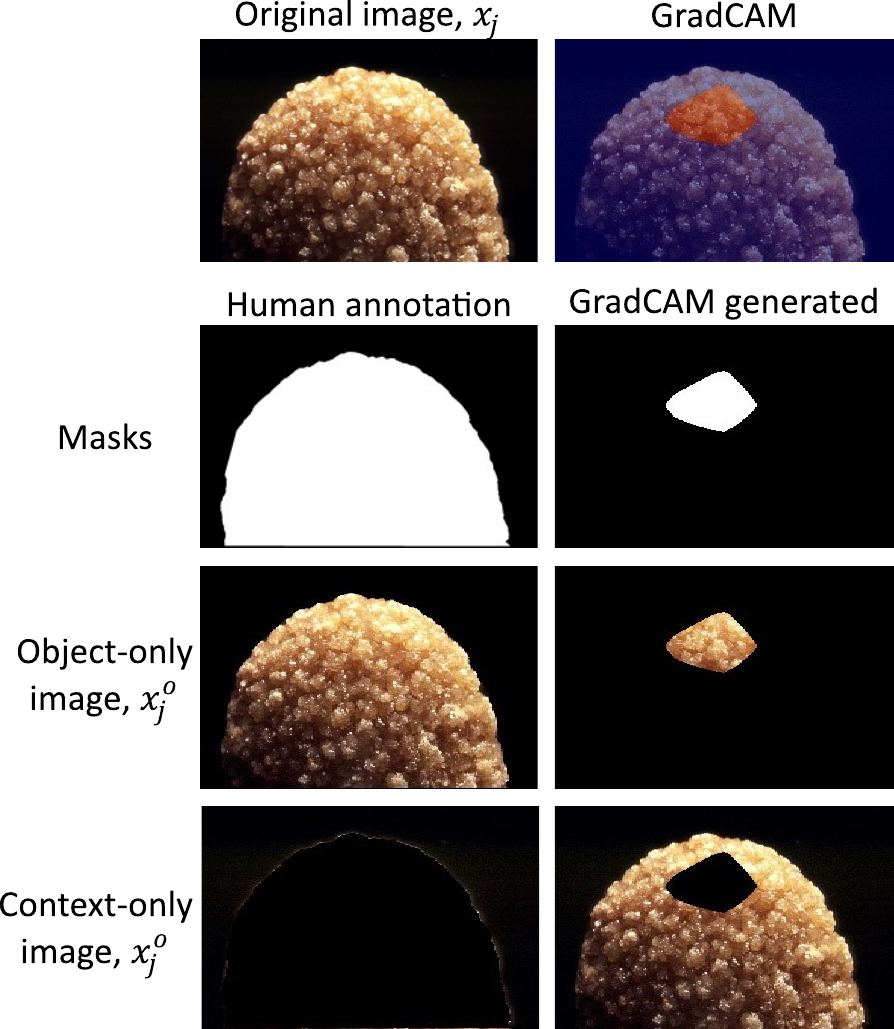}
    \caption{\small{Top left: an example image $x_j$ from the dataset. Top right: GradCAM for the corresponding predicted class, thresholded to keep 30\% of the highest values, in red.
    From the human-annotated segmentations of the dataset and the thresholded GadCAM segmentations, the ``Object-only" $x_j^o$ and ``Context-only" $x_j^c$ cutout images were obtained.}} 
\label{fig_grad_CAM_KS_2}
\end{figure*}

\subsection{Method: Causal Explanation Score (CaES)}

Inspired by \cite{Causal_signals_LopezPaz_2017_CVPR}, we modified slightly their proposal ``Neural Causation Coefficient (NCC)" \cite{Causal_signals_LopezPaz_2017_CVPR} for measuring causal/anti-causal scores. 
Our modification is the transformation of the original NCC scores to be bounded between 0 and 1, as seen in Eq.\ref{eq_caes_scores}, to ease their comparison and readability.
The NCC network is a binary classifier that indicates if a feature activation from the last convolutional layer of a CNN is considered to be causally or anti-causally related to the output of the model. 
To establish a reliable baseline in this work and for future implementations, we employ ResNet18, a Convolutional Neural Network that has been extensively utilized in the field as the classifier on which the NCC model will evaluate causal scores. 
This classification is given for each of the 512 feature maps, $f_l\in\mathbb{R}^{512}$, since this is output size from the residual blocks of the ResNet18 $f$, considered to be the feature extractor of the model.

The NCC architecture is trained following the previous work \cite{Causal_signals_LopezPaz_2017_CVPR}, using the hyper-parameters and configuration from the implementation in \cite{euphoria2023_NCC}. 
The dataset used for the test was the T\"ubingen dataset, version 1.0 \cite{Joris_2016_Distinguishing_Cause}, which is a collection of a hundred observational cause-effect samples of the real world.
The T\"ubingen dataset is commonly utilized as a standard reference in the causal inference field.
The trained NCC model obtained a 72\% instead of the previously reported 79\% accuracy when classifying causal and anti-causal relationships. Although this difference highlights an area of improvement, is useful for our current aim of showing if a causal score is more accurate when the `object of interest' is identified by an explainable method, as opposed to being annotated by humans.

The feature maps obtained from the $m$ input images $x_j$ and output $y_j$, per class $k \in \{1, \dots, 6\}$, are processed as input pairs set $(X,Y)$ by the NCC classifier. 
In this way, we obtain NCC outputs for each feature map of each input image and average them per feature map.
For each category $k$ and each feature $f_l$, we must therefore determine whether feature $f_l$ is likely to be an object feature $f_l^o$  or a context feature $f_l^c$.
To that effect, we prepared two alternate versions of each input image $x_j$, the ``Object-only" $x_j^o$ and ``Context-only" $x_j^c$ cutouts.
The ``Object-only" $x_j^o$, indicated by the white area of the segmentation mask, denotes the section of the input image that contains the ``object" corresponding to the label of the image $x_j$.
Complementary, the ``Context-only" $x_j^c$, is the black area in the segmentation mask and indicates the ``context" or background in the input image $x_j$. 
An example of both, ``Object-only" $x_j^o$ and the ``Context-only" $x_j^c$ images, for both types of segmentation (Human-annotated and GradCAM) can be seen in Fig.\ref{fig_grad_CAM_KS_2}.

The highest 1\% causal and anti-causal scores, for the averaged feature maps per class $k$, are selected to report their \textit{feature ratios} accordingly to Eq.\ref{eq_feature_scores}.
Those \textit{feature ratios} allow us to determine how much each feature $f_l$ is imputable to the segmentation mask of the object (object-feature ratio, $s_j^o$) of category $k$, or the context (context-feature ratio, $s_j^c$).

\begin{equation}
\label{eq_feature_scores}
s_j^o  = \frac{\sum_{j=1}^m{|}f_{jl}^c - f_{jl}{|}}{\sum_{j=1}^m{|}f_{jl}{|}} , \ \ \ \ 
s_j^c  = \frac{\sum_{j=1}^m{|}f_{jl}^o - f_{jl}{|}}{\sum_{j=1}^m{|}f_{jl}{|}} ,
\end{equation}

\textbf{NCC adaptation: }In this proposal, the previous ``Object" $s_j^o$ and ``Context" $s_j^c$ scores are adapted to be transformed from positive values to be bounded between 0 and 1, in the following Eq.\ref{eq_caes_scores}:

\begin{equation}
\label{eq_caes_scores}
\sigma_o(s_j^o) = \frac{2}{1 + e^{-s_j^o}}- 1 , \ \ \ \     
\sigma_c(s_j^c) = \frac{2}{1 + e^{-s_j^c}}- 1 ,
\end{equation}

The ResNet18 used as the classifier, was trained on the kidney stone dataset described at Sec.\ref{KS_dataset}.
The training consisted of 30 epochs, using Adam optimizer, with a learning rate of 0.0001. 
This ResNet18 has two fully connected layers with 512 neurons and the final output layer has 6 neurons for the classification of the 6 classes of kidney stones.
From this ResNet18 classifier model $f$, its convolutional layers were used as the feature extractor.

\textbf{Human-annotated segmentation masks: }
object-only and context-only images of the kidney stones were obtained from human-annotated segmentation masks of 0 and 1 values. 
The object-only is obtained, by multiplying the original input image and its corresponding segmentation mask.
Then we subtract from the original image the object-only image, resulting in the context-only image.

\textbf{Segmentation masks with Grad-CAM: }
GradCAM is characterized by calculating a heatmap, from the activations of a DL model and its gradients.
However, GradCAM explanations can produce only zero values for some inputs, then we apply the modification of element-wise squaring the GradCAM saliency matrix prior to the activation step with the ReLU. 
This is intended to keep in the GradCAM heatmap the most salient values, independently of their original sign.

GradCAM maps indicate the most relevant area in the input image for its corresponding classification \cite{GradCAM_Ramprasaath_2016}. 
On those heatmaps a threshold was used, to retain the highest 30\% activations in the heatmap (the most important area) as the Object-only portion, with values of 1, and the 70\% least important area as the Context-only portion, with values of 0. In this way, we obtained segmentation masks from GradCAM.
This process is repeated for all images in the dataset, giving us a set of GradCAM-generated segmentations. Finally, the same process for the ``Human-annotated segmentation masks" is applied using the GradCAM-generated segmentations to obtain their corresponding ``object-only" and ``Context-only" images, an example of the results obtained can be seen in Fig.\ref{fig_grad_CAM_KS_2}.


The \textit{NCC adaptation}, which is the modification to the \textit{feature ratios} ``$s$" in Eq.\ref{eq_caes_scores}, and the usage of automatically obtained segmentation masks from the adapted GradCAM, is our proposed  Causal Explanation Score (CaES) method.
\begin{figure*}[t]
    \centering
    \captionsetup[subfigure]{justification=centering} 
    
    \begin{subfigure}[b]{0.48\textwidth}   
        \centering
        \includegraphics[width=1.0\textwidth]{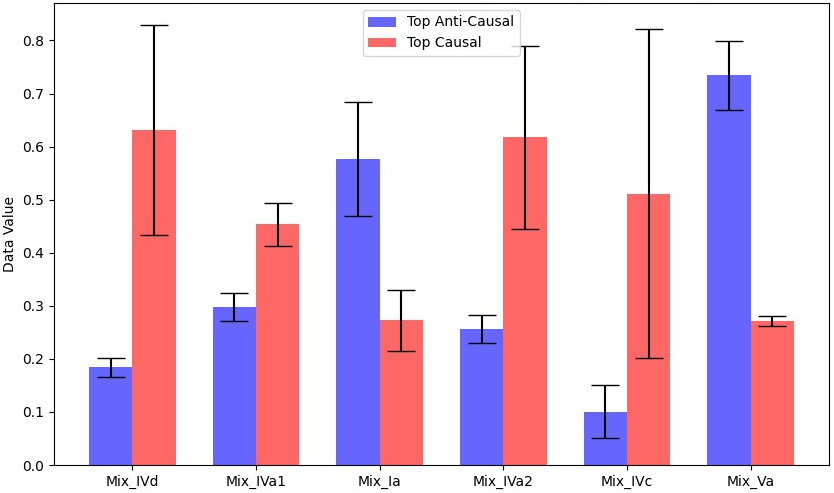}   
        \caption{``Object-only" CaES with human-annotated masks}
        \label{fig_CaES_results_a}  
    \end{subfigure}
    \begin{subfigure}[b]{0.48\textwidth}   
        \centering
        \includegraphics[width=1.0\textwidth]{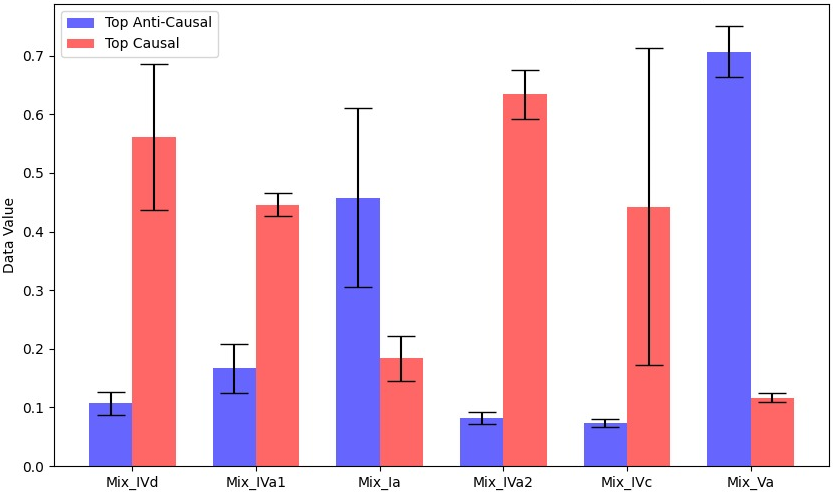}   
        \caption{``Object-only" CaES with the adapted Grad-CAM masks}
        \label{fig_CaES_results_b}  
    \end{subfigure}
    \caption{CaES (Causal Explanation Score) for the ``Object-only" cutouts. In \ref{fig_CaES_results_a} human-annotated masks were used and in \ref{fig_CaES_results_b} masks were obtained from the adapted Grad-CAM method. In both cases the CaES score was obtained for the ``object".}
    \label{fig_CaES_results}  
\end{figure*}

\section{Results and Discussion}
\begin{figure*}[t]
    \centering
    \captionsetup[subfigure]{justification=centering} 
    
    \begin{subfigure}[b]{0.48\textwidth}   
        \centering
        \includegraphics[width=1.0\textwidth]{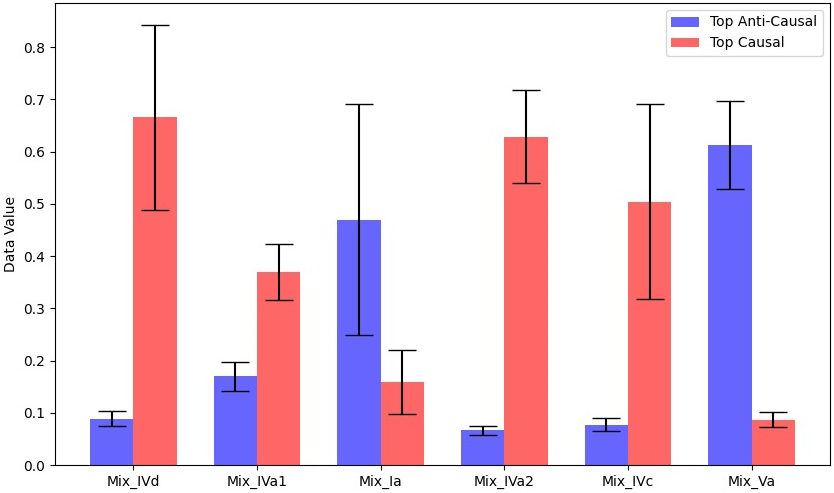}   
        \caption{``Context-only" CaES with human-annotated masks}
        \label{fig_CaES_results_context_a}  
    \end{subfigure}
    \begin{subfigure}[b]{0.48\textwidth}   
        \centering
        \includegraphics[width=1.0\textwidth]{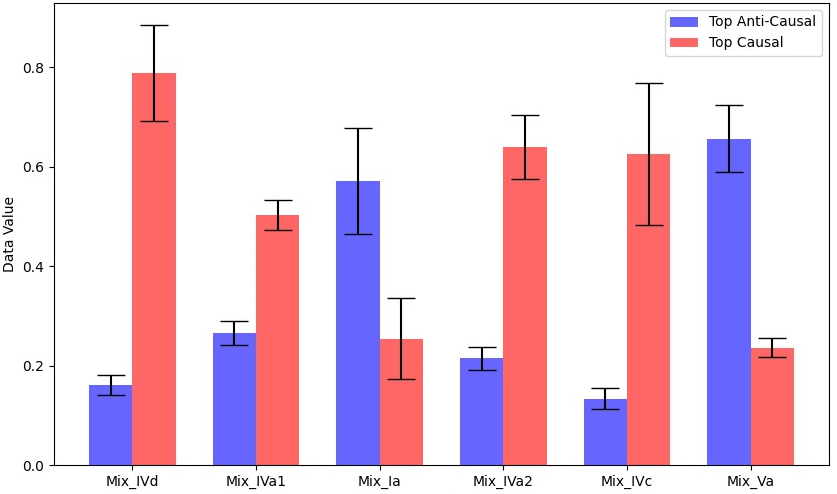}   
        \caption{``Context-only" CaES with the adapted Grad-CAM masks}
        \label{fig_CaES_results_context_b}  
    \end{subfigure}
    \caption{CaES (Causal Explanation Score) for the ``Context-only" cutouts. In \ref{fig_CaES_results_context_a} human-annotated masks were used and in \ref{fig_CaES_results_context_b} masks were obtained from the adapted Grad-CAM method. In both cases the CaES score was obtained for the ``context".}
    \label{fig_CaES_results_context}  
\end{figure*}

As seen in Fig.\ref{fig_CaES_results} and Fig.\ref{fig_CaES_results_context}, the causal/anti-causal measurements based on human-annotated masks and heatmap explanations are possible, even with less variance than with time-consuming human-annotated segmentation masks. 
Also, the results obtained between the human-annotated segmentation masks Fig.\ref{fig_CaES_results_a} and the results from the generated masks with Grad-CAM segmentation, Fig.\ref{fig_CaES_results_b} were remarkably similar in the values of their averages, also for the ``Context-only" results in Fig.\ref{fig_CaES_results_context_a} and Fig.\ref{fig_CaES_results_context_b}. 
GradCAM segmentation mask results present the benefit of a lower variance than the results from manual annotations that cut out the complete stone for the ``Object-only" scores.

\textbf{Limitations:}
Nor the previous work used as inspiration \cite{Causal_signals_LopezPaz_2017_CVPR}, nor our CaES proposal so far consider the case for identification of confounded relationships between pair of variables.
Additional measurements need to be carried out with different Convolutional Neural Networks (CNNs) to analyze how consistent the results are.
Using an arbitrary threshold value could be limiting the results obtained with the GradCAM segmentation masks.
A key finding is that for most classes, 4 out of 6, the ``Object-only" score in Fig.\ref{fig_CaES_results} the causal score are predominant, contradictorily to the finding in \cite{Causal_signals_LopezPaz_2017_CVPR}.
This key difference may be due to the lower NCC performance score for the classification of causal signals, and the limited amount of data samples in the dataset, of only 366 for all 6 classes.

\section{Conclusions and Future Directions}
Causal measurements based on the most relevant features in the input image are favorable. 
Our method, CaES, shows it's possible to automate the causal measurements by having access to the DL model weights.
Furthermore, with our proposal, CaES, it is possible to give to the specialists an indication of which features of a model are the most relevant and if those hold a causal or anti-causal relationship with the output.
Nonetheless, further experiments are needed.

As a future direction, different levels of thresholds should be explored to identify the optimal value for assessing both causal and anti-causal scores.
For our dataset, in particular, this is relevant, due to the original big areas of black background and the segmentation masks obtained from GradCAM being empirically small, as observed in Fig.\ref{fig_grad_CAM_KS_2}.
An interesting direction to explore for improvements is modifying the ``Object" and ``Context" scores to be based on a metric learning approach on the latent space extracted by the feature extractor instead of the activation change of the original image against the ``Object" or ``Context" only cutouts.
Finally, implementing different XAI methods to obtain the segmentation masks is left out for future work.

\section*{Acknowledgments}

The authors wish to acknowledge the Mexican Council for Science and Technology (CONAHCYT) for the support in terms of postgraduate scholarships in this project, and the Data Science Hub at Tecnologico de Monterrey for their support on this project. 
This work has been supported by Azure Sponsorship credits granted by Microsoft's AI for Good Research Lab through the AI for Health program. The project was also supported by the French-Mexican ANUIES CONAHCYT Ecos Nord grant 322537.

\section*{Compliance with ethical approval}
The images were captured in medical procedures following the ethical principles outlined in the Helsinki Declaration of 1975, as revised in 2000, with the consent of the patients.

\bibliographystyle{splncs04}
\bibliography{egbib}

\end{document}